\documentclass[letterpaper, 10 pt, conference]{ieeeconf}  

\IEEEoverridecommandlockouts                              

\usepackage{graphicx}
\usepackage{kotex}
\usepackage{tabularx}
\usepackage{adjustbox}
\usepackage{url}
\usepackage{color}
\usepackage{cite}
\usepackage{array}
\usepackage[table]{xcolor}
\usepackage{booktabs}
\usepackage{bbding}
\usepackage{pifont}
\usepackage{amssymb}
\usepackage{tabularray}
\usepackage{colortbl}
\usepackage{caption}
\usepackage{multirow}
\usepackage{nicematrix}
\usepackage{float}

\title{\LARGE \bf
From General Vision to Reliable Traversability Estimation: Adapting Vision Foundation Models for Unstructured Outdoor Environments}

\author{Ji-Hoon Hwang, Jisung Bae, Dong-Wook Kim, Yeonkyu Lee and Seung-Woo Seo}
\begin{document}
\captionsetup[table]{position=bottom}
\captionsetup[figure]{skip=3pt}

\maketitle
\thispagestyle{empty}
\pagestyle{empty}

\begin{abstract}
Vision-based approaches have become the dominant paradigm for traversability estimation in unstructured outdoor environments, typically adapting vision foundation models (VFMs) via semantic segmentation supervision. 
However, this paradigm faces three fundamental challenges that undermine its reliability: the task-agnostic design of VFMs, the ambiguity of traversability annotations, and the discrepancy between semantic labels and physical safety. 
We propose Vision-to-Traversability Adaptation (ViTA), a framework that adapts VFMs for reliable traversability estimation, instantiated on SAM2.
ViTA injects task-specific knowledge through learnable traversability prompts 
while preserving the VFM's cross-domain generalization.
To handle annotation ambiguity, we introduce Perspective-Diversified Training, 
which estimates semantic uncertainty to suppress confident predictions at ambiguous boundaries.
To bridge the semantic-traversability discrepancy, 
we distill geometric knowledge during training, 
enabling slope and elevation reasoning from RGB images alone at inference.
The semantic and geometric outputs are fused into a continuous traversability score that reflects both semantic uncertainty and geometric risk. 
Evaluations across diverse domains, including challenging real-world off-road datasets, demonstrate that ViTA achieves state-of-the-art IoU and Precision with substantial false-positive reduction and strong cross-domain generalization.
\end{abstract}

 
\section{INTRODUCTION}

Autonomous navigation in unstructured outdoor environments, 
such as off-road trails, remains one of the most demanding frontiers of mobile robotics. 
Unlike structured scenes with explicit cues such as lane markings and curbs, 
these environments exhibit continuously shifting appearance 
and terrain varying across soil, vegetation, and slopes. 
Hazards are easily mistaken for traversable regions, 
and a single false positive can immobilize the robot. 
Reliable traversability estimation, the task of confidently determining 
where a robot can safely traverse, is therefore a safety-critical capability 
for autonomous navigation in unstructured terrain.

\begin{figure}[t!]
\centering
\includegraphics[width=0.93\linewidth]{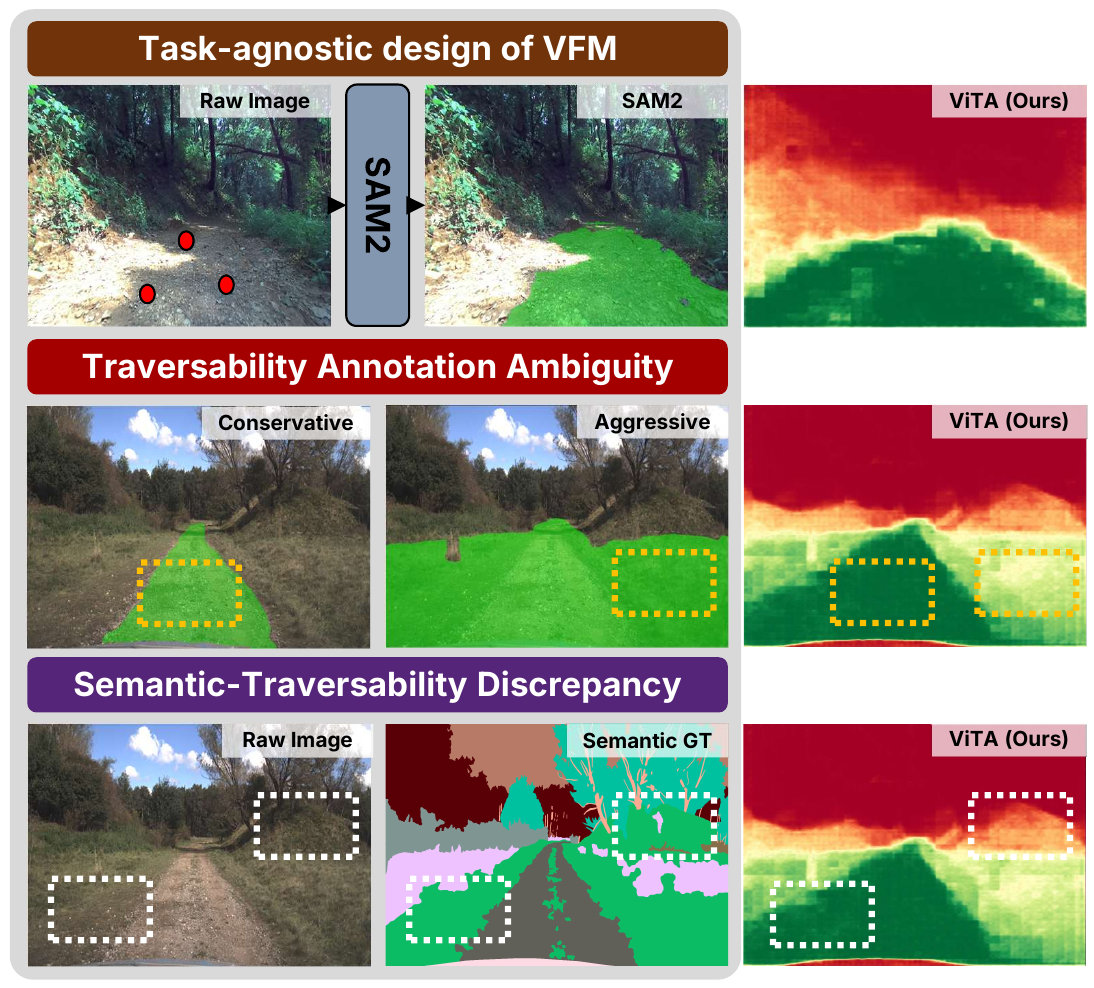}
\captionsetup{font=small}
\caption{\textbf{Fundamental challenges in adapting Vision Foundation Models 
for traversability estimation in unstructured outdoor environments.} 
Top: SAM2 segments task-agnostically based on spatial correspondence, splitting traversable terrain along superficial visual cues (e.g., illumination boundaries). 
Middle: traversability annotations vary substantially across annotators and datasets (orange boxes). 
Bottom: semantic labels ignore geometric risks such as slope and elevation (white boxes). 
The right column shows ViTA's continuous traversability scores that address each challenge.}
\label{introfig}
\vspace{-0.4cm}
\end{figure}

Ideally, traversability is inferred by fusing visual appearance with active depth sensing, 
but SWaP constraints and the cost of large-scale paired RGB-depth data~\cite{swap} 
make vision-only approaches the dominant paradigm~\cite{hwang2025stseg}. 
To obtain visual representations that generalize across diverse environments, 
Vision Foundation Models (VFMs) have emerged as the visual backbone, 
whose large-scale pretraining provides strong cross-domain generalization. 
Building on this, adapting VFMs via semantic segmentation labels 
that remap traversability-relevant classes such as road and terrain as positive labels 
has become the standard approach.
While effective in structured scenes, this paradigm faces 
three fundamental challenges that undermine its reliability in unstructured environments:

\textit{1) Task-Agnostic Design of Vision Foundation Models.}
VFMs are optimized for general visual objectives rather than physical traversability: 
DINOv3~\cite{dinov3} with self-supervised representations 
and SAM2~\cite{sam2} with prompt-based spatial correspondence 
lack understanding of traversability~\cite{vfmunderstand} (Fig.~\ref{introfig}, top). 
Such task-agnostic representations are acceptable in structured scenes, 
where visual boundaries naturally align with traversable regions. 
In unstructured environments, however, this absence becomes a critical weakness: 
unclear boundaries and shifting visual conditions cause the model 
to segment along superficial cues. 
Reliable traversability therefore requires injecting task-specific knowledge 
while preserving the VFM's cross-domain generalization.

\textit{2) Traversability Annotation Ambiguity.}
In unstructured terrain, traversable boundaries are inherently ambiguous, transitioning gradually rather than at a sharp line. 
Without explicit guidelines tied to a specific robot platform, annotation becomes inherently subjective, 
yielding substantially different traversable labels across annotators and datasets (Fig.~\ref{introfig}, middle). 
A model supervised with such inconsistent labels amplifies the false-positive risk~\cite{noise}. 
Rather than collapsing this inherent ambiguity into a single prediction, 
reliable traversability requires modeling prediction uncertainty 
to suppress confident predictions at ambiguous boundaries.

\textit{3) Semantic-Traversability Discrepancy.}
In structured scenes, road and terrain classes largely correspond to flat navigable surfaces, 
making semantic labels a close proxy for physical safety. 
In unstructured terrain, however, a single semantic class often spans geometrically heterogeneous regions:
semantically safe terrain such as grass may lie on a steep slope 
or be elevated above the navigable ground plane (Fig.~\ref{introfig}, bottom). 
Although VFMs implicitly capture some geometric structure, 
this knowledge alone cannot disambiguate such cases. 
Reliable traversability therefore requires explicit reasoning 
over both semantic appearance and geometric risks such as slope and elevation.

To address these challenges, we propose 
Vision-to-Traversability Adaptation (ViTA), a framework 
that adapts prompt-based Vision Foundation Models
for reliable traversability estimation in unstructured outdoor environments, instantiated on SAM2.
To inject task-specific knowledge while preserving the VFM's generalization, 
we introduce Learnable Traversability Prompts jointly supervised by semantic and geometric cues, 
eliminating prompt dependency at inference. 
To handle annotation ambiguity, we propose Perspective-Diversified Training (PDT), 
whose structurally diverse predictions yield a reliable semantic uncertainty signal. 
To bridge the semantic-traversability discrepancy, we distill geometric knowledge from a foundation depth model 
into slope and elevation risk maps during training, enabling geometry-aware estimation from RGB alone at inference. 
The three outputs are fused into a continuous traversability score 
that yields reliable traversability estimation from a single RGB image 
and naturally supports platform-adaptive downstream planning.

The main contributions are summarized as follows:
\begin{itemize}
\item We propose ViTA, a VFM adaptation framework (instantiated on SAM2) 
that jointly reasons over semantic uncertainty and geometric risk 
for reliable traversability estimation in unstructured outdoor environments.

\item We propose Perspective-Diversified Training (PDT), a multi-hypothesis training strategy that enables 
inter-token variance as a reliable epistemic uncertainty proxy, 
substantially suppressing false-positive predictions without explicit uncertainty supervision.

\item We validate ViTA across diverse unseen domains including public urban and off-road datasets, 
as well as real-robot test sets directly collected with a mobile robot, 
demonstrating state-of-the-art IoU and Precision with substantial false-positive reduction compared to all baselines.
\end{itemize}

\section{Related Works}

\begin{figure*}[t!]
      \centering
      \includegraphics[width=1\linewidth]{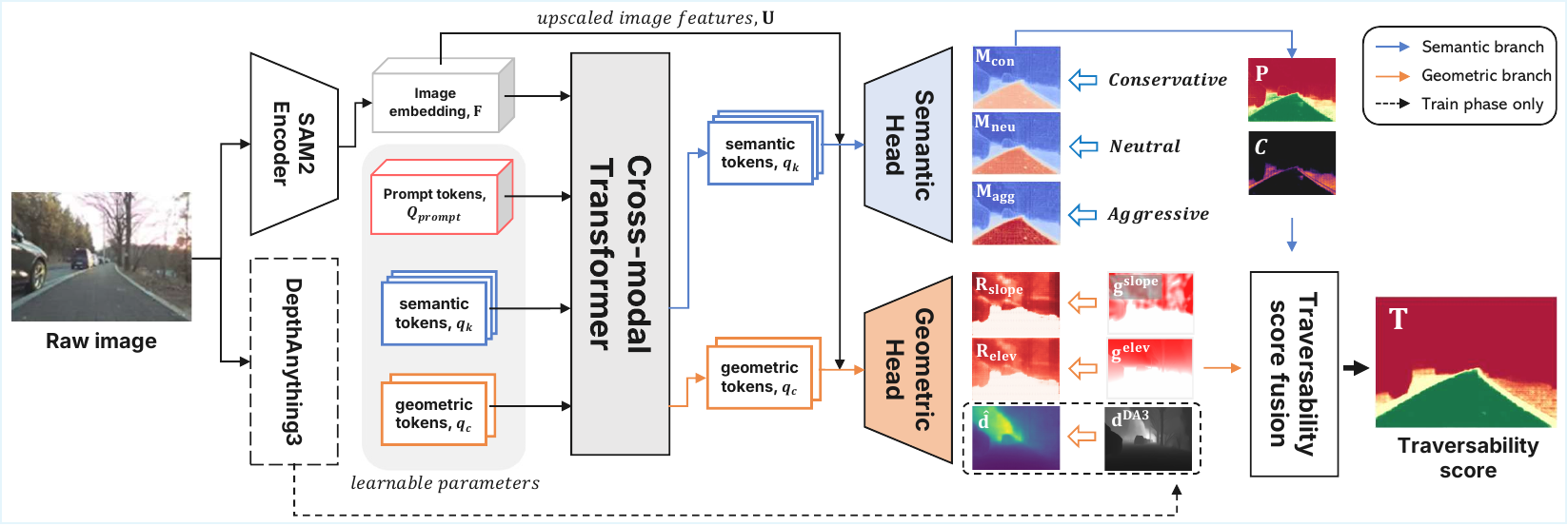}
      \captionsetup{font=small}
    \caption{Overview of the ViTA framework.
      The SAM2 encoder extracts dense image features,
      which are jointly processed with Learnable Traversability Prompt Tokens,
      Semantic Mask Tokens, and Geometric Tokens
      through a Cross-Modal Transformer.
      The Semantic Head decodes three perspective-diversified masks
      ($\mathbf{M}_{\text{con}}, \mathbf{M}_{\text{neu}}, \mathbf{M}_{\text{agg}}$)
      to estimate a mean semantic traversability map $\mathbf{P}$
      and an inter-token variance map $\mathbf{p}_{\text{var}}$ as an epistemic uncertainty proxy.
      The Geometric Head predicts slope and elevation risk maps (\(\mathbf{R}_{\text{slope}}, \mathbf{R}_{\text{elev}}\))
      supervised by pseudo-labels from a frozen DepthAnything3 teacher (training only).
      Both outputs are fused into a continuous traversability score
      via conservative probability multiplication.}
      \label{fig:overview}
      \vspace{-0.4cm}
   \end{figure*}
   
\subsection{Vision-based Traversability Estimation}
Vision-based methods emerged as an alternative to LiDAR-based 
approaches~\cite{swap}, with adapting visual backbones via semantic 
segmentation supervision becoming the standard. 
GA-Nav~\cite{guan2022ganav} and ST-Seg~\cite{hwang2025stseg} define 
traversability-relevant terrain classes on datasets such as 
GOOSE~\cite{goose}.
However, semantic labels provide only a coarse proxy for physical safety, 
failing to capture geometric risks such as slope and elevation in unstructured terrain.
Self-supervised methods~\cite{jung2024vstrong, stepp} 
bypass manual annotation by leveraging traversed trajectories 
or proprioceptive feedback as supervision, 
but remain limited to the distribution of previously encountered terrains 
and struggle to assess negative regions without direct robot experience.
Recent VFM-based works adapt VFMs as the visual backbone for traversability estimation:
DINOv3~\cite{dinov3} is used as a general-purpose feature extractor,
and GeNIE~\cite{wang2025genie} fine-tunes 
SAM2 with a learnable prompt token for diverse environments. 
However, existing VFM-based methods either rely on task-agnostic 
representations that lack explicit traversability knowledge, 
or address only prompt dependency without injecting task-specific understanding 
into the learned features.

\subsection{Uncertainty Estimation in Dense Prediction}
 
Modeling annotation ambiguity in dense prediction 
is commonly addressed through uncertainty estimation.
Uncertainty in deep learning is broadly categorized into
aleatoric uncertainty from irreducible data noise
and epistemic uncertainty from model knowledge limitations~\cite{kiureghian2009aleatory}.
In dense prediction, the former typically concentrates at image edges 
and cannot capture the perspective-level ambiguity 
inherent in traversability labels.
While classical approaches such as MC Dropout~\cite{gal2016dropout} 
and Deep Ensembles incur significant computational overhead, Multiple 
Choice Learning (MCL)~\cite{lee2016stochastic} offers a lightweight 
alternative via Winner-Takes-All (WTA) training, with inter-hypothesis 
variance serving as an epistemic uncertainty proxy~\cite{ilg2018uncertainty}.
However, WTA-based methods suffer from hypothesis collapse.
Soft variants such as aMCL~\cite{amcl} mitigate collapse
by distributing gradients across all hypotheses,
but because all tokens share identical loss functions,
predictions converge and variance loses its meaning
as an uncertainty measure.
ViTA addresses this through PDT,
which assigns each token a distinct perspective
on traversability conservativeness via asymmetric loss weighting,
structurally guaranteeing prediction diversity.

\subsection{Geometric Information in Vision-based Navigation}

Classical methods fuse LiDAR-derived elevation maps
with semantic segmentation for traversability,
but require depth sensors at inference.
The DepthAnything series~\cite{yang2024depthanything, depth3}
established strong monocular depth foundation models,
demonstrating that rich geometric knowledge
can be extracted from RGB images alone.
However, existing methods either require depth input at inference~\cite{orfd}
or treat geometric reasoning as a post-processing step
decoupled from the primary task~\cite{zeng2025navidiffusor}.
ViTA distills geometric knowledge from a frozen depth foundation model
exclusively during training
and bidirectionally refines semantic and geometric tokens
through a shared transformer,
enabling each modality to inform the other
before producing traversability estimates.

 
\section{METHODOLOGY}
 
\subsection{Problem Formulation}
 
Given a single RGB image $\mathbf{I} \in \mathbb{R}^{H \times W \times 3}$,
the goal is to predict a continuous traversability map 
$\mathbf{T} \in [0,1]^{H \times W}$,
where higher values indicate greater traversability confidence.
Our framework produces four pixel-wise outputs:
a semantic traversability map 
$\mathbf{P} \in [0,1]^{H \times W}$,
an uncertainty-based confidence weight 
$\mathbf{C} \in [0,1]^{H \times W}$,
and geometric risk maps 
$\mathbf{R}_{\text{slope}}, \mathbf{R}_{\text{elev}} \in [0,1]^{H \times W}$:
\begin{equation}
  \mathbf{P},\; \mathbf{C},\; \mathbf{R}_{\text{slope}},\; \mathbf{R}_{\text{elev}} = f_\theta(\mathbf{I}),
\end{equation}
where $f_\theta$ is optimized on a semantic segmentation dataset 
$\mathcal{D}_{\text{train}}$ and expected to generalize to unseen 
domains $\mathcal{D}_{\text{test}}$. The semantic branch is supervised 
by binary labels remapped from semantic annotations 
(traversability-relevant classes as positive), while the geometric 
branch uses pseudo-labels from a foundation depth model during training only.

\subsection{Overview}
 
As illustrated in Fig.~\ref{fig:overview},
the framework adapts SAM2 with three core components, chosen for its 
mask decoder with multi-mask tokens that align with our prompt-based 
and multi-hypothesis mechanisms.
The SAM2 image encoder extracts dense features
$\mathbf{F} \in \mathbb{R}^{h \times w \times C}$
from the input image.
The Cross-Modal Transformer decoder
takes $\mathbf{F}$ as key/value inputs
and operates on a unified sequence of learnable query tokens
comprising Traversability Prompt Tokens,
Semantic Mask Tokens, and Geometric Tokens (Section~\ref{sec:cmt}).
The refined tokens are processed by the Semantic Head (Section~\ref{sec:semantic}) and the Geometric Head (Section~\ref{sec:geometric}), whose outputs are fused into the final traversability score $\mathbf{T}$ (Section~\ref{sec:fusion}).

\subsection{Cross-Modal Transformer}
\label{sec:cmt}
 
\subsubsection{Token Design}
 
The Cross-Modal Transformer takes dense image features $\mathbf{F}$
as key/value inputs for cross-attention,
and a unified sequence of learnable query tokens.
 
\textit{Learnable Traversability Prompt Tokens.}
SAM2 requires explicit point or box prompts at inference,
presupposing prior knowledge of target regions.
We replace these with Learnable Traversability Prompt Tokens
$\mathbf{Q}_{\text{prompt}} \in \mathbb{R}^{N_p \times D}$,
randomly initialized and jointly optimized with semantic and geometric supervision to encode task-specific traversability priors.
 
\textit{Semantic Mask Tokens.}
Three independently parameterized Semantic Mask Tokens
$\{q_{\text{con}}, q_{\text{neu}}, q_{\text{agg}}\} \in \mathbb{R}^{D}$,
which repurpose multi-mask tokens of SAM2, 
serve as a multi-hypothesis set for traversability prediction.
Each token is assigned a distinct perspective
on the conservativeness spectrum
and decoded through a dedicated hypernetwork MLP
to produce a candidate traversability mask (Section~\ref{sec:semantic}).
 
\textit{Geometric Tokens.}
Two tokens $q_{\text{slope}}$ and $q_{\text{elev}}$
predict physical terrain risks,
each decoded through a dedicated hypernetwork MLP (Section~\ref{sec:geometric}).
 
\subsubsection{Bidirectional Cross-Modal Interaction}

The Cross-Modal Transformer adopts the 2-way transformer decoder
of SAM2~\cite{sam2},
which alternates between token-to-token self-attention
and token-to-image cross-attention within each block.
We refer the reader to~\cite{sam2} for full architectural details.
The key modification in ViTA is that all semantic and geometric tokens 
are concatenated into a single query sequence, 
so that joint self-attention naturally enables 
bidirectional information flow between modalities. 
In particular, $\mathbf{Q}_{\text{prompt}}$ receives gradient signals 
from both supervision sources, 
producing prompt representations that encode 
a unified understanding of traversability.

\subsection{Perspective-Diversified Training}
\label{sec:semantic}
 
\subsubsection{Multi-Mask Prediction}
 
Each refined Semantic Mask Token is decoded through a hypernetwork MLP, 
followed by an inner product with the upscaled image features $\mathbf{U}$ at each spatial location, where $\mathbf{U}$ is obtained by upsampling the transformer output image features:
\begin{equation}
  \mathbf{M}_k = \text{MLP}_k(q_k) \cdot \mathbf{U} \in \mathbb{R}^{H \times W},
  \quad k \in \{\text{con}, \text{neu}, \text{agg}\}.
\end{equation}
 
\subsubsection{PDT Loss Formulation}
 
\begin{figure}[t]
      \centering
      \includegraphics[width=1\linewidth]{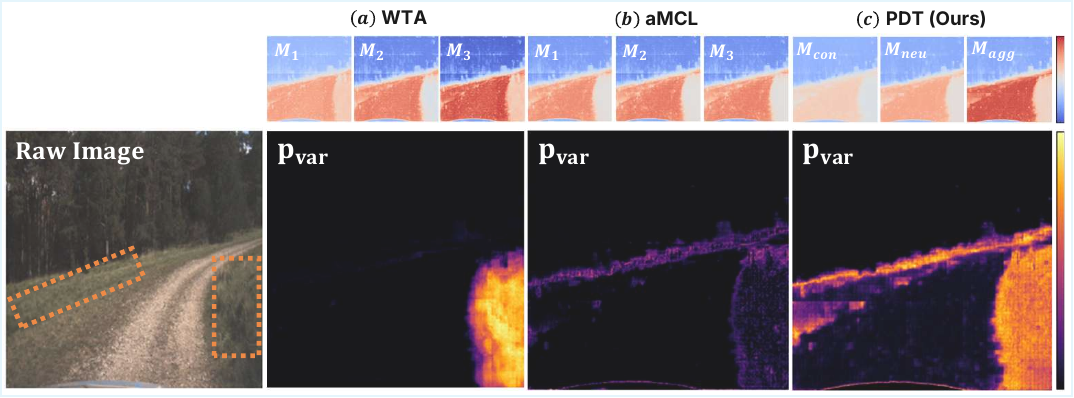}
      \captionsetup{font=small}
        \caption{Comparison of multi-mask predictions and uncertainty maps 
        across training strategies. 
        (a) Standard WTA trains each token without explicit guidance 
        on how to specialize, resulting in unconstrained predictions 
        and unreliable uncertainty. 
        (b) aMCL distributes gradients across all tokens via softmin weighting, 
        but tokens converge to similar predictions, 
        yielding near-zero variance. 
        (c) PDT (Ours) assigns distinct conservativeness perspectives, 
        producing structurally diverse predictions and meaningful uncertainty 
        at ambiguous terrain boundaries.}
      \label{fig:pdt}
   \end{figure}

We observe that traversability annotation ambiguity
is fundamentally a question of \textit{conservativeness}:
how aggressively should ambiguous terrain boundaries
be classified as traversable?
Rather than relying on training dynamics to induce diversity,
we explicitly assign each token a distinct perspective
on this conservativeness spectrum
through asymmetric loss weighting (Fig.~\ref{fig:pdt}).
Each token $k$ is supervised with a perspective-specific loss
composed of a focal loss~\cite{focal} with perspective weighting
parameter $\gamma_k$ and a Tversky loss~\cite{tversky} with asymmetric 
FP/FN penalty weights $\alpha^{\text{FP}}_k$, $\alpha^{\text{FN}}_k$:
\begin{equation}
  \mathcal{L}_k = 
    \mathcal{L}_{\text{focal}}(\mathbf{M}_k, \mathbf{Y};\, \gamma_k)
    + \mathcal{L}_{\text{Tversky}}(\mathbf{M}_k, \mathbf{Y};\,
    \alpha^{\text{FP}}_k, \alpha^{\text{FN}}_k)
\end{equation}
where $\mathbf{Y} \in \{0,1\}^{H \times W}$ is the binary label from semantic annotations.
The Tversky loss~\cite{tversky} generalizes the Dice loss
by independently weighting false positive and false negative contributions:
\begin{equation}
  \mathcal{L}_{\text{Tversky}} = 1 - 
  \frac{\text{TP} + \epsilon}
  {\text{TP} + \alpha^{\text{FP}}_k \cdot \text{FP} 
  + \alpha^{\text{FN}}_k \cdot \text{FN} + \epsilon}
\end{equation}
where $\text{TP}$, $\text{FP}$, and $\text{FN}$ denote 
the pixel-wise true positive, false positive, and false negative counts
between $\sigma(\mathbf{M}_k)$ and $\mathbf{Y}$,
and $\epsilon$ is a small constant for numerical stability.
A high $\alpha^{\text{FP}}_k$ penalizes predicting untraversable regions 
as traversable, and vice versa for $\alpha^{\text{FN}}_k$.
Because focal loss governs per-pixel classification difficulty
while Tversky loss governs region-level FP/FN balance,
applying both asymmetrically in the same conservativeness direction
provides stronger reinforcement than either term alone.
The conservative token receives low $\gamma_{\text{con}}$
and high $\alpha^{\text{FP}}_{\text{con}}$,
jointly suppressing false positives
so that only clearly safe terrain is predicted as traversable.
The aggressive token receives high $\gamma_{\text{agg}}$
and high $\alpha^{\text{FN}}_{\text{agg}}$,
jointly suppressing false negatives
to broadly cover any potentially traversable region.
The neutral token uses balanced weights
to faithfully follow the ground truth.
This formulation provides a structural guarantee of prediction diversity:
even under identical ground truth labels,
each token interprets the annotation
through its assigned perspective on traversability risk (Fig.~\ref{fig:pdt}c).
The total semantic loss is:
\begin{equation}
  \mathcal{L}_{\text{sem}} = \mathcal{L}_{\text{con}} + \mathcal{L}_{\text{neu}} + \mathcal{L}_{\text{agg}} .
\end{equation}
 
\subsubsection{Uncertainty Estimation}
 
At inference, inter-token disagreement directly reflects
the ambiguity of traversability at each pixel.
Regions where all three tokens agree
(clearly safe or clearly hazardous) produce low variance,
while ambiguous boundaries
where conservative and aggressive tokens disagree
produce high variance.
The per-pixel mean \(\mathbf{P}\) and variance \(\mathbf{p}_{\text{var}}\) are:
\begin{equation}
\small
  \mathbf{P} = \frac{1}{N_s}\sum_{k}\sigma(\mathbf{M}_k),
  \quad
  \mathbf{p}_{\text{var}} = \frac{1}{N_s}\sum_{k}\bigl(\sigma(\mathbf{M}_k) - \mathbf{P}\bigr)^2
\end{equation}
where $N_s=3$ is the number of tokens
and $\sigma$ denotes the sigmoid function.
The variance is normalized by its per-image maximum
and converted into a pixel-wise confidence weight
$\mathbf{C} \in [0,1]^{H \times W}$:
\begin{equation}
  \tilde{\mathbf{p}}_{\text{var}} = \frac{\mathbf{p}_{\text{var}}}{\max(\mathbf{p}_{\text{var}})+\epsilon},
  \quad
  \mathbf{C} = 1 - \alpha \cdot \tilde{\mathbf{p}}_{\text{var}}
\end{equation}
where $\alpha$ controls the suppression strength.
At maximum uncertainty ($\tilde{\mathbf{p}}_{\text{var}}{=}1$),
the confidence is reduced to $1{-}\alpha$,
attenuating the traversability score at ambiguous regions.
 
\subsection{Geometric Distillation}
\label{sec:geometric}
 
\subsubsection{Geometric Risk Prediction}
 
The refined Geometric Tokens predict a slope risk map $\mathbf{R}_{\text{slope}} \in [0,1]^{H \times W}$
and an elevation risk map $\mathbf{R}_{\text{elev}} \in [0,1]^{H \times W}$:
\begin{equation}
  \mathbf{R}_c = \sigma\bigl(\text{MLP}_c(q_c) \cdot \mathbf{U}\bigr),
  \quad c \in \{\text{slope}, \text{elev}\}
\end{equation}
 
\subsubsection{Geometric Pseudo-label Generation}
 
Geometric supervision is derived from a frozen 
DepthAnything3~\cite{depth3} teacher
used exclusively during training.
Rather than distilling the depth map directly,
we convert it into slope and elevation risk
through classical surface normal and ground plane estimation,
so that the geometric branch learns
traversability-relevant risk representations
rather than generic depth features.
Given the relative depth map $\mathbf{D} \in \mathbb{R}^{H \times W}$,
a ground reference normal $\mathbf{n}_{\text{gnd}}$ and plane offset $b$
are estimated by SVD fitting over traversable regions
identified by the semantic ground-truth mask.
The slope risk is computed from surface normals 
obtained via central differences:
\begin{equation}
  g^{\text{slope}}_{u,v} = (1 - |\mathbf{n}_{u,v} \cdot \mathbf{n}_{\text{gnd}}|) \cdot \frac{D_{u,v}}{\tilde{D}}
\end{equation}
where $\mathbf{n}_{u,v}$ is the surface normal at pixel $(u,v)$
and $\tilde{D}$ is the median depth.
The depth ratio $D_{u,v}/\tilde{D}$ downweights slope estimates
at distant regions, where monocular relative depth is less reliable
and surface normal computation becomes noisy.
The elevation risk is derived from the perpendicular distance 
of each point above the ground plane:
\begin{equation}
  h_{u,v} = (\mathbf{x}_{u,v} \cdot \mathbf{n}_{\text{gnd}} - b)^+,
  \quad
  g^{\text{elev}}_{u,v} = 1 - \exp(-\beta \cdot h_{u,v})
\end{equation}
where $\mathbf{x}_{u,v} = (u, v, D_{u,v})^\top$ is the 
pseudo-3D coordinate at pixel $(u,v)$, constructed from pixel coordinates and relative depth without camera intrinsics,
$(\cdot)^+$ denotes the ReLU operator 
retaining only points above the ground plane,
and $\beta$ controls the sensitivity of the elevation risk.
 
\subsubsection{Geometric Loss}
The geometric branch is supervised by pseudo risk maps \(\mathbf{g}^{\text{slope}}, \mathbf{g}^{\text{elev}}\) in Smooth L1 loss:
\begin{equation}
\small
  \mathcal{L}_{\text{geo}} = \sum_{c \in \{\text{slope},\, \text{elev}\}} \lambda_c \cdot
    \mathbb{E}\bigl[\text{SmoothL1}\bigl(\mathbf{R}_c,\; \mathbf{g}^c\bigr)\bigr].
\end{equation}
While $\mathcal{L}_{\text{geo}}$ trains the geometric tokens on derived risk representations, 
the Auxiliary Depth Head, a lightweight convolutional decoder
attached to the shared encoder,
directly injects depth knowledge into its feature representations
via a Scale-Shift Invariant (SSI) loss~\cite{ssi}:
\begin{equation}
  \mathcal{L}_{\text{distill}} = \mathbb{E}\bigl[|\hat{s} \cdot \hat{\mathbf{d}} + \hat{t} - \mathbf{d}^{\text{DA3}}|\bigr]
\end{equation}
where $\hat{\mathbf{d}}$ is the predicted relative depth,
$\mathbf{d}^{\text{DA3}}$ is the pseudo-depth from the frozen teacher,
and $\hat{s}$, $\hat{t}$ are obtained by least-squares regression
that minimizes $\|\hat{s} \cdot \hat{\mathbf{d}} + \hat{t} - \mathbf{d}^{\text{DA3}}\|^2$
over all pixels per image.

\subsection{Traversability Score Fusion}
\label{sec:fusion}

The final traversability score is obtained
via conservative probability multiplication:
\begin{equation}
\small
  \mathbf{T} = \mathbf{C} \cdot \mathbf{P} \cdot
    \left(1 - \frac{\mathbf{R}_{\text{slope}} + \mathbf{R}_{\text{elev}}}{2}\right).
\end{equation}
The score is high only when the semantic branch predicts high traversability,
the token ensemble agrees with low uncertainty,
and the geometric risk is small.
A single source of hazard suffices to suppress the overall score,
reflecting the asymmetric cost of false-positive predictions
in safety-critical navigation.
The total training loss is:
\begin{equation}
  \mathcal{L} = \mathcal{L}_{\text{sem}} + \lambda_{\text{geo}}(\mathcal{L}_{\text{geo}} + \mathcal{L}_{\text{distill}}).
\end{equation}

\begin{table*}[t]
\centering
\setlength{\tabcolsep}{3.0pt}
\renewcommand{\arraystretch}{1.00}
\caption{Performance comparison across all target datasets.
Values are reported as IoU / Precision / Recall (\%).
\textit{Type}: SS (semantic segmentation), VFM (vision foundation model-based),
SELF (self-supervised traversability).
FZ: encoder frozen (head only), FT: full fine-tuning, $^\dagger$ denotes self-collected real-robot test sets.}
\begin{adjustbox}{width=\linewidth,center}
\begin{tabularx}{\linewidth}{>{\raggedright\arraybackslash}p{1.9cm} >{\centering\arraybackslash}p{0.7cm} *{6}{>{\centering\arraybackslash}X}}
\toprule
\multicolumn{1}{c}{\multirow{2}{*}[-0.6ex]{Method}} & \multirow{2}{*}[-0.6ex]{Type} & \multicolumn{3}{c}{Unstructured Outdoor} & \multicolumn{2}{c}{Structured Outdoor} & \multicolumn{1}{c}{Appearance Shifts} \\
\cmidrule(lr){3-5} \cmidrule(lr){6-7} \cmidrule(lr){8-8}
 & & GOOSE-val & Campus$^\dagger$ & Mountain$^\dagger$ & Cityscapes & ACDC & GOOSE-val-C \\
\midrule
SegFormer~\cite{segformer} & SS & 72.4 / 73.1 / \underline{98.8} & 77.4 / 78.4 / 98.4 & 73.2 / 75.4 / 96.1 & 78.9 / 80.8 / 97.2 & 64.7 / 65.9 / 97.2 & 60.6 / 66.5 / 87.3 \\
Mask2Former~\cite{mask2former} & SS & 69.5 / 70.0 / \textbf{98.9} & 76.7 / 77.7 / 98.3 & 77.2 / 80.4 / 95.2 & 78.7 / 79.8 / 98.2 & 68.8 / 69.7 / 98.1 & 61.6 / 65.6 / 91.1 \\
GANav~\cite{guan2022ganav} & SS & 68.2 / 71.5 / 93.5 & 53.1 / 76.5 / 63.4 & 61.2 / 80.9 / 71.5 & 71.6 / 76.7 / 91.6 & 53.7 / 62.8 / 78.8 & 47.2 / 67.1 / 61.3 \\
ST-Seg~\cite{hwang2025stseg} & SS & 70.8 / 72.2 / 97.2 & 77.5 / 78.0 / \textbf{99.1} & 77.2 / 78.2 / \underline{98.3} & 79.7 / 80.7 / 98.5 & 69.3 / 69.7 / \textbf{99.2} & 67.5 / 69.1 / \textbf{96.8} \\
\midrule
DINOv3 FZ~\cite{dinov3} & VFM & 68.3 / 72.1 / 92.9 & 43.0 / 63.0 / 57.5 & 77.6 / \underline{84.6} / 90.4 & 59.1 / 82.6 / 67.5 & 48.7 / 50.8 / 92.3 & 40.9 / 62.3 / 54.4 \\
DINOv3 FT~\cite{dinov3} & VFM & 68.4 / 69.0 / 98.6 & 64.6 / 70.4 / 88.7 & 76.4 / 79.7 / 94.9 & 76.6 / 78.3 / 97.3 & 60.3 / 63.6 / 93.8 & 53.0 / 67.3 / 71.4 \\
SAM2 FZ~\cite{sam2} & VFM & 73.8 / 74.5 / 98.7 & 75.4 / 76.0 / \underline{99.0} & 77.7 / 78.3 / \textbf{99.1} & 80.1 / 80.8 / \underline{98.9} & 68.7 / 69.7 / 97.9 & 67.3 / 71.2 / 92.3 \\
SAM2 FT~\cite{sam2} & VFM & 71.0 / 71.7 / 98.7 & 78.7 / 79.5 / 98.8 & 79.3 / 82.9 / 94.8 & 80.7 / 81.2 / \textbf{99.2} & \underline{70.9} / \underline{71.3} / \underline{98.9} & \underline{69.4} / 72.3 / \underline{94.5} \\
GeNIE~\cite{wang2025genie} & VFM & \underline{75.9} / 75.8 / 97.6 & \underline{81.9} / 81.1 / 98.9 & \underline{81.6} / 83.3 / 96.2 & \underline{81.1} / 79.0 / 98.7 & 70.8 / 71.1 / 97.3 & 50.8 / 59.1 / 57.8 \\
\midrule
STEPP~\cite{stepp} & SELF & 71.2 / \underline{76.2} / 91.6 & 75.2 / \underline{81.9} / 90.1 & 79.5 / 82.8 / 98.2 & 59.2 / \underline{83.9} / 66.8 & 36.4 / 66.1 / 44.7 & 56.3 / \underline{72.6} / 71.4 \\
\midrule
\rowcolor{gray!10}
\textbf{ViTA (Ours)} & VFM & \textbf{82.9} / \textbf{90.5} / 89.8 & \textbf{84.7} / \textbf{94.7} / 87.9 & \textbf{87.0} / \textbf{93.3} / 91.8 & \textbf{85.1} / \textbf{85.4} / 95.5 & \textbf{74.7} / \textbf{78.7} / 91.1 & \textbf{69.7} / \textbf{82.6} / 79.0 \\
\bottomrule
\end{tabularx}
\end{adjustbox}
\vspace{-0.5cm}
\label{tab:main_results}
\end{table*}

\section{Results and Analysis}
\subsection{Experimental Setup}
 
\noindent\textbf{Datasets.}
For fair comparison, all methods, including baselines and ViTA, are retrained on the GOOSE training split ($7845$ images)~\cite{goose}, whose mixed off-road and urban scenes provide a suitable basis for cross-domain generalization evaluation.
We remap the original classes into binary traversability labels by assigning road, drivable vegetation, and terrain as positive, and all remaining classes as negative.
For evaluation, we use 7 target domains, with four unstructured outdoor benchmarks directly probing traversability: GOOSE-val ($962$ images), ORFD ($2193$ images)~\cite{orfd}, and two self-collected real-robot test sets, Campus ($449$ images, mixed paved/unpaved university grounds) and Mountain ($847$ images, unstructured mountain trails with steep slopes and dense vegetation).
ORFD provides official binary traversability annotations across diverse off-road scenes and weather conditions, which we use directly; GOOSE-val is relabeled from its semantic annotations, and Campus and Mountain are newly annotated, all following the same protocol of marking the vehicle-traversed path and surrounding passable regions while excluding steep slopes and impassable surfaces.
Cityscapes ($500$ images)~\cite{cityscapes} and ACDC ($1600$ images)~\cite{acdc} additionally verify generalization to structured outdoor scenes using their official road segmentation masks.
To further assess robustness under the appearance shifts characteristic of outdoor environments, GOOSE-val-C applies 15 corruption types at five severity levels to the GOOSE validation set~\cite{corruption}.
All evaluation domains except GOOSE-val are entirely unseen during training.
We will publicly release all test images and ground truth masks.
 
\noindent\textbf{Evaluation Protocol.}
Although ViTA produces a continuous traversability score, for quantitative comparison we threshold at $\tau=0.5$ and evaluate all methods with three binary metrics:
\begin{equation}
\small
\text{IoU} = \frac{\text{TP}}{\text{TP}+\text{FP}+\text{FN}},\;\;
\text{Prec.} = \frac{\text{TP}}{\text{TP}+\text{FP}},\;\;
\text{Rec.} = \frac{\text{TP}}{\text{TP}+\text{FN}}
\end{equation}
In the traversability context, Precision measures the reliability of traversable predictions: low Precision implies the robot is frequently directed into hazardous terrain (false positives). Recall measures coverage of traversable regions: low Recall yields overly conservative but still safe paths (false negatives). Because false positives carry far greater risk than false negatives in safety-critical navigation, Precision is the primary safety-relevant metric. The full continuous score is presented in the qualitative results.
 
\subsection{Implementation Details}
We build ViTA on SAM2-Base+ with an input resolution of 
$512 \times 512$. The token dimensionality is $D = 256$,
and the Cross-Modal Transformer follows the SAM2 
two-way decoder configuration.
The number of Learnable Traversability 
Prompt Tokens is set to $N_p = 4$. 
For PDT, the perspective weighting parameters are 
$\gamma_k \in \{0.2, 0.5, 0.8\}$ and the Tversky weights are 
$(\alpha^{\text{FP}}_k, \alpha^{\text{FN}}_k) \in 
\{(3.0,\, 0.3),\; (1.0,\, 1.0),\; (0.3,\, 3.0)\}$ 
for the conservative, neutral, and aggressive tokens, 
respectively. The uncertainty suppression strength is 
$\alpha = 0.7$. The elevation risk sensitivity is 
$\beta = 3$. The total loss combines semantic and geometric 
terms with $\lambda_{\text{geo}} = 2$, as the geometric branch 
requires stronger supervision when initialized from SAM2 
pretrained weights. Slope and elevation losses are weighted 
equally ($\lambda_{\text{slope}} = \lambda_{\text{elev}} = 1$). 
The geometric teacher is DepthAnything 
3-Large~\cite{depth3}.
The Auxiliary Depth Head is a lightweight convolutional decoder
attached to the shared encoder, used only during training.
All methods are trained for 10 epochs using AdamW and a cosine annealing scheduler. We apply component-wise learning rates: $5 \times 10^{-6}$ 
for the Cross-Modal Transformer and semantic branch, 
$1 \times 10^{-5}$ for the image encoder, and 
$5 \times 10^{-4}$ for the geometric branch. ViTA is fully 
fine-tuned and data 
augmentation follows the default SAM2 protocol. All 
experiments are conducted on a single NVIDIA RTX 3090 GPU.


\subsection{Results and Analysis}
 
\noindent\textbf{Baselines.}
We compare against three categories: (1)~semantic segmentation models (SegFormer~\cite{segformer}, Mask2Former~\cite{mask2former}, GA-Nav~\cite{guan2022ganav}, ST-Seg~\cite{hwang2025stseg}), (2)~VFM-based methods (DINOv3~\cite{dinov3}, SAM2~\cite{sam2}, GeNIE~\cite{wang2025genie}) with frozen (Fz) and fine-tuned (FT) encoders, and (3)~a self-supervised method (STEPP~\cite{stepp}). The backbone and model size of each method are detailed in Section~\ref{sec:cost}.

\noindent\textbf{Cross-Domain Performance.}
Table~\ref{tab:main_results} reports results across six domains grouped into 
unstructured outdoor, structured outdoor, and appearance-shift benchmarks.
ViTA achieves the highest IoU in every domain.
On the unstructured outdoor benchmarks, ViTA reaches 82.9\% IoU on GOOSE-val (+7.0pp over GeNIE), 
84.7\% on Campus, and 87.0\% on Mountain, 
demonstrating reliable traversability estimation under the ambiguous boundaries 
and geometric heterogeneity characteristic of unstructured terrain.
This unstructured-focused design further generalizes to structured outdoor scenes, 
achieving 85.1\% and 74.7\% IoU on Cityscapes and ACDC respectively, 
demonstrating that task-specific adaptation preserves cross-domain generalization.
SAM2 (FZ/FT) consistently outperforms DINOv3 (FZ/FT) across all domains, 
confirming its suitability for pixel-wise dense prediction and supporting 
our backbone choice; ViTA further improves over SAM2 FT by +11.9pp on GOOSE-val, 
indicating that the additional gain stems from our adaptation strategy 
rather than the backbone alone.

\noindent\textbf{Robustness under Appearance Shifts.}
The appearance-shift benchmark GOOSE-val-C applies 15 image corruption types~\cite{corruption} to GOOSE-val, where ViTA attains 69.7\% IoU, substantially outperforming GeNIE (50.8\%) and surpassing ST-Seg (67.5\%), the latter explicitly designed to mitigate distribution shift.
This confirms that ViTA inherits SAM2's robustness to continuous 
appearance shifts while delivering substantially higher accuracy.

\begin{figure*}[t]
\centering
\includegraphics[width=0.97\textwidth]{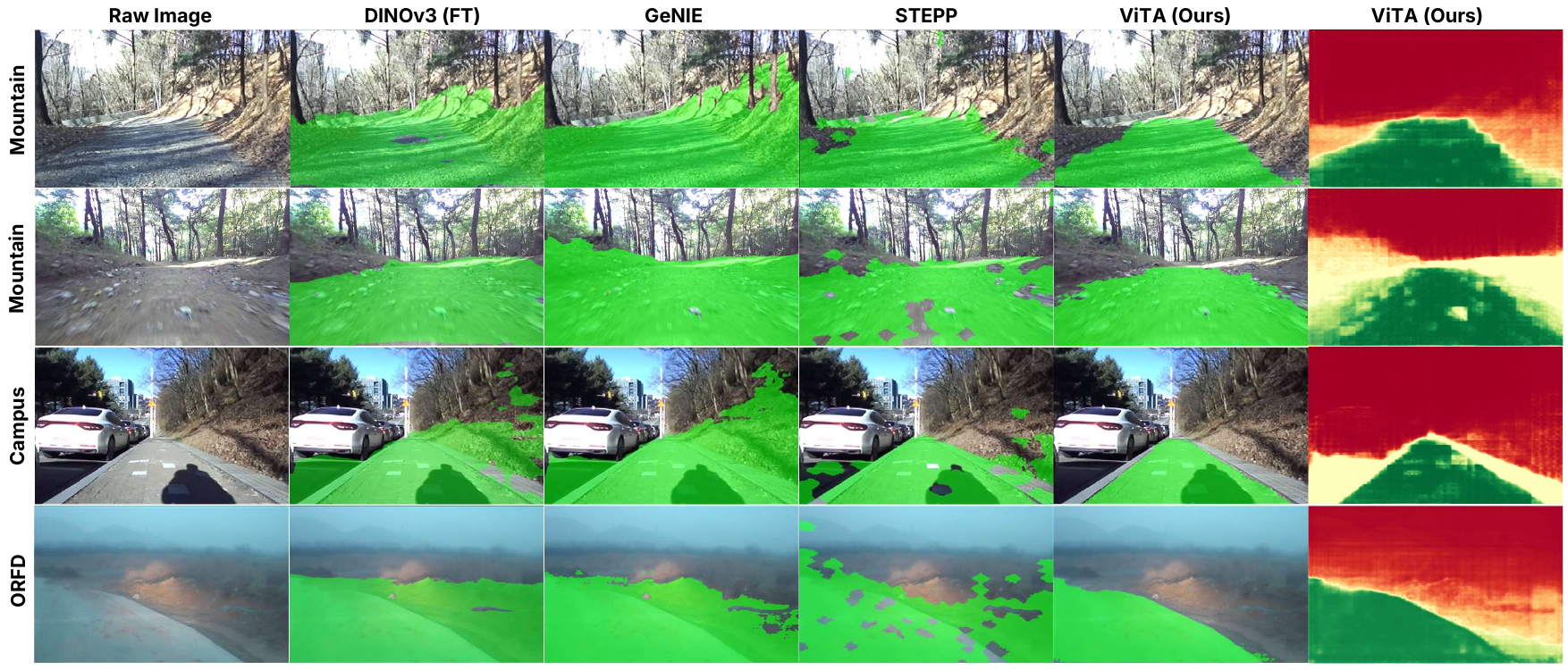}
\captionsetup{font=small}
\caption{Qualitative comparison across evaluation domains. Columns 2--4 show binary masks from representative baselines, column 5 shows the ViTA binary mask, and column 6 shows the ViTA continuous traversability score (green: high, red: low).}
\label{fig:qualitative}
\vspace{-0.5cm}
\end{figure*}

\noindent\textbf{Precision--Recall Trade-off.}
ViTA exhibits a pronounced Precision gain with moderate Recall reduction.
On GOOSE-val, Precision reaches 90.5\% versus GeNIE's 75.8\% (+14.7pp), 
while Recall decreases from 97.6\% to 89.8\%.
This trade-off is a natural consequence of ViTA's reliability-focused design: 
suppressing false positives at ambiguous boundaries and geometric hazards 
inevitably tightens predictions, reducing Recall.
Baseline methods, particularly VFM-based models, lack mechanisms for 
uncertainty suppression or geometric risk awareness, 
and therefore produce overly permissive predictions that inflate Recall 
at the cost of false-positive explosion.
For example, SAM2 FT achieves 98.7\% Recall on GOOSE-val but only 71.7\% Precision.
Importantly, the Recall reduction is an artifact of binary thresholding: 
ViTA assigns ambiguous boundaries intermediate continuous scores, 
which the $\tau=0.5$ threshold counts as negative (Fig.~\ref{fig:qualitative}).
Downstream planners that directly consume the continuous traversability score as a costmap 
would be unaffected by this thresholding effect.

\begin{figure*}[t]
\centering
\includegraphics[width=0.95\textwidth]{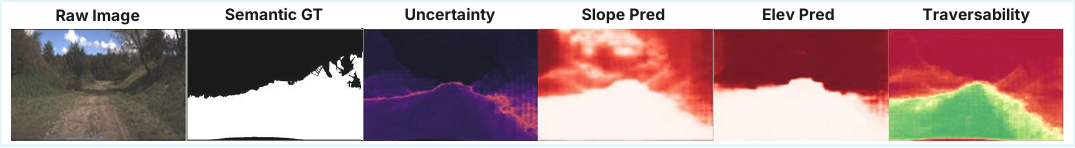}
\captionsetup{font=small}
\caption{ViTA outputs on a GOOSE-val scene where the semantic GT 
labels the hillside as traversable (white). 
From left: raw image, semantic GT, uncertainty, 
predicted slope risk, predicted elevation risk, 
and final traversability score.}
\label{fig:geometric_vis}
\end{figure*}

\begin{table}[t]
\centering
\footnotesize
\setlength{\tabcolsep}{4pt}
\renewcommand{\arraystretch}{1.05}
\newcolumntype{C}{>{\centering\arraybackslash}X}
\caption{Performance comparison on the ORFD test set. 
Supervised methods are trained on the ORFD training set; 
Zero-shot methods are trained on GOOSE training split.}
\begin{tabularx}{\columnwidth}{l|cc|ccc}
\toprule
Method & Supervision & Modality & IoU & Precision & Recall \\
\midrule
OFFNet-RGB~\cite{orfd} & Supervised & RGB & 67.5 & 86.4 & 80.6 \\
OFFNet-SN~\cite{orfd} & Supervised & SN & 78.9 & 83.7 & \textbf{93.2} \\
OFFNet~\cite{orfd} & Supervised & RGB + SN & \textbf{82.3} & \textbf{94.3} & 90.3 \\
\midrule
Mask2Former~\cite{mask2former} & Zero-shot & RGB & 52.2 & 52.8 & 97.9\\
DINOv3 FT~\cite{dinov3} & Zero-shot & RGB & 48.4 & 52.7 & 85.5\\
SAM2 FT~\cite{sam2} & Zero-shot & RGB & 51.7 & 51.8 & \textbf{99.8}\\
STEPP~\cite{stepp} & Zero-shot & RGB & 52.0 & 55.5 & 89.0\\
GeNIE~\cite{wang2025genie} & Zero-shot & RGB & 62.9 & 63.1 & 99.4 \\
\rowcolor{gray!10}
\textbf{ViTA (Ours)} & Zero-shot & RGB & \textbf{74.2} & \textbf{76.6} & 97.2 \\
\bottomrule
\end{tabularx}
\vspace{-0.5cm}
\label{tab:orfd}
\end{table}

\noindent\textbf{Comparison with Supervised and Multi-Modal Methods.}
Table~\ref{tab:orfd} compares zero-shot methods 
against supervised baselines trained on the ORFD training set. 
Among RGB-only methods, 
ViTA achieves 74.2\% IoU in a zero-shot setting, 
surpassing even the fully supervised OFFNet-RGB (67.5\%) 
without observing any ORFD training samples. 
Only OFFNet variants that additionally take 
LiDAR-derived surface normals as input exceed ViTA, 
which is expected given direct access to 3D geometric measurements. 
This result demonstrates that training-only geometric distillation can recover most of the benefit of explicit 3D measurements, supporting the design choice of RGB-only inference.

\noindent\textbf{Qualitative Results.}
Fig.~\ref{fig:geometric_vis} illustrates how ViTA handles the semantic-traversability discrepancy: the semantic GT labels the hillside as fully traversable, yet the predicted slope and elevation risks correctly identify the geometric hazard, and the final score suppresses the region accordingly.
Fig.~\ref{fig:qualitative} presents qualitative comparisons across 
three unseen outdoor domains.
On Mountain (rows 1--2), baselines classify steep slopes as traversable based on semantic texture alone, while ViTA restricts predictions to confidently safe regions through geometric risk and semantic uncertainty, and the continuous score shows a clear green-to-red transition along the slope.
On Campus (row 3), two complementary mechanisms suppress non-target regions: slope risk suppresses the sloped area on the right, while semantic uncertainty reduces the score on the road beneath the curb on the left, where traversability is ambiguous relative to the sidewalk the robot occupies.
On ORFD (row 4), despite dense fog in this scene and a significant domain gap from GOOSE training, ViTA produces a clean road mask with sharp boundary attenuation.
The supplementary video provides additional qualitative results on 
an unstructured mountain driving sequence not included in our 
quantitative evaluation.

\subsection{Ablation Study}
\label{sec:ablation}

\noindent\textbf{Component Analysis.}
Table~\ref{tab:ablation} reports IoU, Precision, and Recall
averaged over all benchmark domains
as modules are progressively added to the base configuration.
A1 reproduces GeNIE~\cite{wang2025genie} 
under identical settings.
Adding Semantic Uncertainty alone (A2)
improves Precision by 10.5pp over A1
with a 2.6pp Recall decrease,
confirming that PDT suppresses false positives
at ambiguous boundaries.
Adding Geometric Distillation alone (A3)
improves Precision by 4.8pp with virtually no Recall loss,
showing complementary suppression at geometrically hazardous regions.
Combining both (A4) yields a 10.3pp Precision gain
with only 0.9pp Recall decrease.
Enabling the Cross-Modal Transformer (A5)
further improves IoU (+1.2pp) and Precision (+1.5pp),
confirming that bidirectional token refinement
benefits both modalities.
The full ViTA model achieves the best IoU (78.52\%), confirming that all components contribute complementarily to reliable traversability estimation.

\begin{table}[!t]
\centering
\footnotesize
\setlength{\tabcolsep}{4pt}
\renewcommand{\arraystretch}{1.05}
\newcolumntype{C}{>{\centering\arraybackslash}X}
\caption{Ablation study with progressive module addition. LP: Learnable Prompt, SU: Semantic Uncertainty, GD: Geometric Distillation, CMT: Cross-Modal Transformer.}
\begin{tabularx}{\columnwidth}{@{}l|cccc|CCC@{}}
\toprule
ID & LP & SU & GD & CMT & IoU & Precision & Recall \\
\midrule
A1 (GeNIE)~\cite{wang2025genie} & \Checkmark &  &  &  & 71.26 & 73.04 & \textbf{89.45} \\
A2 & \Checkmark & \Checkmark & &  & 76.89 & 83.55 & 86.81 \\
A3 & \Checkmark & & \Checkmark &  & 73.31 & 77.87 & 89.42 \\
A4 & \Checkmark & \Checkmark & \Checkmark &  & 77.34 & 83.33 & 88.56 \\
\rowcolor{gray!10}
\textbf{A5 (ViTA)} & \Checkmark & \Checkmark & \Checkmark & \Checkmark & \textbf{78.52} & \textbf{84.80} & 88.87 \\
\bottomrule
\end{tabularx}
\vspace{-0.4cm}
\label{tab:ablation}
\end{table}

\begin{table}[!t]
\centering
\small
\setlength{\tabcolsep}{6pt}
\renewcommand{\arraystretch}{1.05}
\caption{Model complexity comparison. 
GFLOPs measured at $512 \times 512$ input resolution.}
\begin{tabular}{@{}llcc@{}}
\toprule
Method & Backbone & Params(M) & GFLOPs(G) \\
\midrule
SegFormer~\cite{segformer} & MiT-B5 & 84.60 & 109.77  \\
Mask2Former~\cite{mask2former} & ResNet-101 & 62.90 & 86.41\\
GA-Nav~\cite{guan2022ganav} & MiT-B5 & 84.88 & 79.47 \\
ST-Seg~\cite{hwang2025stseg} & MiT-B5 & 84.60 & 109.77 \\
\midrule
DINOv3~\cite{dinov3} & ViT-B & 85.75 & 87.87 \\
SAM2~\cite{sam2} & Hiera-B+ & 69.14 & 72.27 \\
GeNIE~\cite{wang2025genie} & Hiera-B+ & 72.92 & 73.14 \\
\midrule
STEPP~\cite{stepp} & ViT-S & 22.34 & 59.57 \\
\midrule
\textbf{ViTA (Ours)} & Hiera-B+ & 73.20 & 73.58 \\
\bottomrule
\end{tabular}
\label{tab:cost}
\vspace{-0.5cm}
\end{table}

\noindent\textbf{PDT Hyperparameter Sensitivity.}
We varied both the focal loss parameter
$\gamma_k$ and the Tversky asymmetry weights
$(\alpha^{\text{FP}}_{\text{con}},\, \alpha^{\text{FN}}_{\text{agg}})$
across multiple configurations and evaluated on ORFD.
All configurations yielded IoU within 0.7pp of each other
(74.08--74.71\%),
indicating that performance is robust
to the specific choice of perspective weighting parameters.
This demonstrates that the key mechanism of PDT
lies in encouraging structurally diverse perspectives across tokens,
rather than precise calibration of individual loss weights.

\subsection{Computational Cost}
\label{sec:cost}
Table~\ref{tab:cost} reports model parameters and 
computational cost across all compared methods.
ViTA adds only 4.06M parameters and 1.31 GFLOPs over the SAM2-Base+ model,
confirming that the proposed modules introduce 
minimal computational overhead.
On an NVIDIA Jetson AGX Orin, ViTA achieves approximately 9 FPS
at $512 \times 512$ resolution,
supporting practical deployment
for low-speed autonomous navigation in unstructured environments.

\section{CONCLUSIONS}

We presented Vision-to-Traversability Adaptation (ViTA), a VFM adaptation framework instantiated on SAM2
for reliable traversability estimation in unstructured outdoor environments.
Its three core components are Learnable Traversability Prompts, 
Perspective-Diversified Training, and training-only Geometric Distillation, 
which address the task-agnostic design of VFMs, annotation ambiguity, 
and the semantic-traversability discrepancy, respectively,
enabling geometric reasoning from a single RGB image at inference.
Experiments across diverse domains confirm state-of-the-art 
IoU and Precision with substantial false-positive reduction and strong cross-domain generalization.
As future work, the current conservative probability multiplication used for fusion 
could be replaced with a learned strategy supervised end-to-end, 
leveraging real-world driving feedback such as proprioceptive signals 
or traversed trajectory logs for more expressive, platform-adaptive fusion.








\bibliographystyle{ieeetr}
\bibliography{ref}

\end{document}